\documentclass[10pt,twocolumn,letterpaper]{article}
\usepackage{cvpr}
\usepackage[pagebackref=true,breaklinks=true,letterpaper=true,colorlinks,bookmarks=false]{hyperref}
\usepackage{url}            
\usepackage{booktabs}       
\usepackage{amsfonts}       
\usepackage{nicefrac}       
\usepackage{microtype}      
\usepackage{natbib}
\usepackage{graphicx}
\usepackage{wrapfig}
\usepackage{color}

\cvprfinalcopy 

\def\cvprPaperID{9569} 

\ifcvprfinal\fi
\begin{document}
	
\title{Improving the robustness of ImageNet classifiers using elements of human visual cognition}

\author{Emin Orhan\\
	Center for Data Science\\
	New York University\\
	{\tt\small eo41@nyu.edu}
	\and
	Brenden M. Lake\\
	Center for Data Science, Department of Psychology\\
	New York Univeristy\\
	{\tt\small brenden@nyu.edu}
}
	
\maketitle

\begin{abstract}
 We investigate the robustness properties of image recognition models equipped with two features inspired by human vision, an explicit episodic memory and a shape bias, at the ImageNet scale. As reported in previous work, we show that an explicit episodic memory improves the robustness of image recognition models against small-norm adversarial perturbations under some threat models. It does not, however, improve the robustness against more natural, and typically larger, perturbations. Learning more robust features during training appears to be necessary for robustness in this second sense. We show that features derived from a model that was encouraged to learn global, shape-based representations \citep{geirhos2019} do not only improve the robustness against natural perturbations, but when used in conjunction with an episodic memory, they also provide additional robustness against adversarial perturbations. Finally, we address three important design choices for the episodic memory: memory size, dimensionality of the memories and the retrieval method. We show that to make the episodic memory more compact, it is preferable to reduce the number of memories by clustering them, instead of reducing their dimensionality.
\end{abstract}

\section{Introduction}
ImageNet-trained deep neural networks (DNNs) are state of the art models for a range of computer vision tasks and are currently also the best models of the human visual system and primate visual systems more generally \citep{schrimpf2018}. Yet, they have serious deficiencies as models of human and primate visual systems: 1) they are extremely sensitive to small adversarial perturbations imperceptible to the human eye \citep{szegedy2013}, 2) they are much more sensitive than humans to larger, more natural perturbations \citep{geirhos2018}, 3) they rely heavily on local texture information in making their predictions, whereas humans rely much more on global shape information \citep{geirhos2019, brendel2019}, 4) a fine-grained, image-by-image analysis suggests that images that ImageNet-trained DNNs find hard to recognize do not match well with the images that humans find hard to recognize \citep{rajalingham2018}. 

Here, we add a fifth under-appreciated deficiency: 5) human visual recognition has a strong episodic component lacking in DNNs. When we recognize a coffee mug, for instance, we do not just recognize it as \textit{a} mug, but as \textit{this particular} mug that we have seen before or as a novel mug that we have not seen before. This sense of familiarity/novelty comes automatically, involuntarily, even when we are not explicitly trying to judge the familiarity/novelty of an object we are seeing. More controlled psychological experiments also confirm this observation: humans have a phenomenally good long-term recognition memory with a massive capacity even in difficult one-shot settings \citep{standing1973, brady2008}. Standard deep vision models, on the other hand, cannot perform this kind of familiarity/novelty computation naturally or automatically, since this information is available to a trained model only indirectly and implicitly in its parameters. 

What does it take to address these deficiencies and what are the potential benefits, if any, of doing so other than making the models more human-like in their behavior? In this paper, we address these questions. We show that a minimal model incorporating an explicit key-value based episodic memory does not only make it psychologically more realistic, but also reduces the sensitivity to small adversarial perturbations. It does not, however, reduce the sensitivity to larger, more natural perturbations and it does not address the heavy local texture reliance issue. In the episodic memory, using features from DNNs that were trained to learn more global shape-based representations \citep{geirhos2019} addresses these remaining issues and moreover provides additional robustness against adversarial perturbations. Together, these results suggest that two basic ideas motivated and inspired by human vision, a strong episodic memory and a shape bias, can make image recognition models more robust to both natural and adversarial perturbations at the ImageNet scale.

\section{Related work}
In this section, we review previous work most closely related to ours and summarize our own contributions.  

To our knowledge, the idea of using an episodic cache memory to improve the adversarial robustness of image classifiers was first proposed in \citet{zhao2018} and in \citet{papernot2018}. \citet{zhao2018} considered a differentiable memory that was trained end-to-end with the rest of the model. This makes their model computationally much more expensive than the cache models considered here, where the cache uses pre-trained features instead. The deep \textit{k}-nearest neighbor model in \citet{papernot2018} and the ``CacheOnly'' model described in \citet{orhan2018} are closer to our cache models in this respect, however these works did not consider models at the ImageNet scale. More recently, \citet{dubey2019} did consider cache models at the ImageNet scale (and beyond) and demonstrated substantial improvements in adversarial robustness under certain threat models. 

None of these earlier papers addressed the important problem of robustness to natural perturbations and they did not investigate the effects of various cache design choices, such as the retrieval method (i.e. a continuous cache vs. nearest neighbor retrieval), cache size, dimensionality of the keys or the feature type used (e.g. texture-based vs. shape-based features), on the robustness properties of the cache model. 

A different line of recent work addressed the question of robustness to natural perturbations in ImageNet-trained DNNs. In well-controlled psychophysical experiments with human subjects, \citet{geirhos2018} compared the sensitivity of humans and ImageNet-trained DNNs to several different types of natural distortions and perturbations, such as changes in contrast, color or spatial frequency content of images, image rotations etc. They found that ImageNet-trained DNNs are much more sensitive to such perturbations than human subjects. More recently, \citet{hendrycks2019} introduced the ImageNet-C and ImageNet-P benchmarks to measure the robustness of neural networks against some common perturbations and corruptions that are likely to occur in the real world. We use the ImageNet-C benchmark below to measure the robustness of different models against natural perturbations.

This second line of work, however, did not address the question of adversarial robustness. An adequate model of the human visual system should be robust to both natural and adversarial perturbations.\footnote{Two recent papers \citep{elsayed2018, zhou2019} suggested that humans might be vulnerable, or at least sensitive, to adversarial perturbations too. However, these results apply only in very limited experimental settings (e.g. very short viewing times in \citet{elsayed2018}) and require relatively large and transferable perturbations, which often tend to yield meaningful features resembling the target class.} Moreover, both properties are clearly desirable properties in practical image recognition systems, independent of their value in building more adequate models of the human visual system.

Our main contributions in this paper are as follows: 1) as reported in previous work \citep{zhao2018, papernot2018, orhan2018, dubey2019}, we show that an explicit cache memory improves the adversarial robustness of image recognition models at the ImageNet scale, but only under certain threat scenarios; 2) we investigate the effects of various design choices for the cache memory, such as the retrieval method, cache size, dimensionality of the keys and the feature type used for extracting the keys; 3) we show that caching, by itself, does not improve the robustness of classifiers against natural perturbations; 4) using more global, shape-based features \citep{geirhos2019} in the cache does not only improve robustness against natural perturbations, but also provides extra robustness against adversarial perturbations as well.\footnote{Code for reproducing the results is available at: \url{https://github.com/eminorhan/robust-vision}}

\section{Methods}
\subsection{Models}
\begin{figure}
	\centering
	\includegraphics[width=0.3\textwidth,trim=0mm 0mm 0mm 0mm,clip]{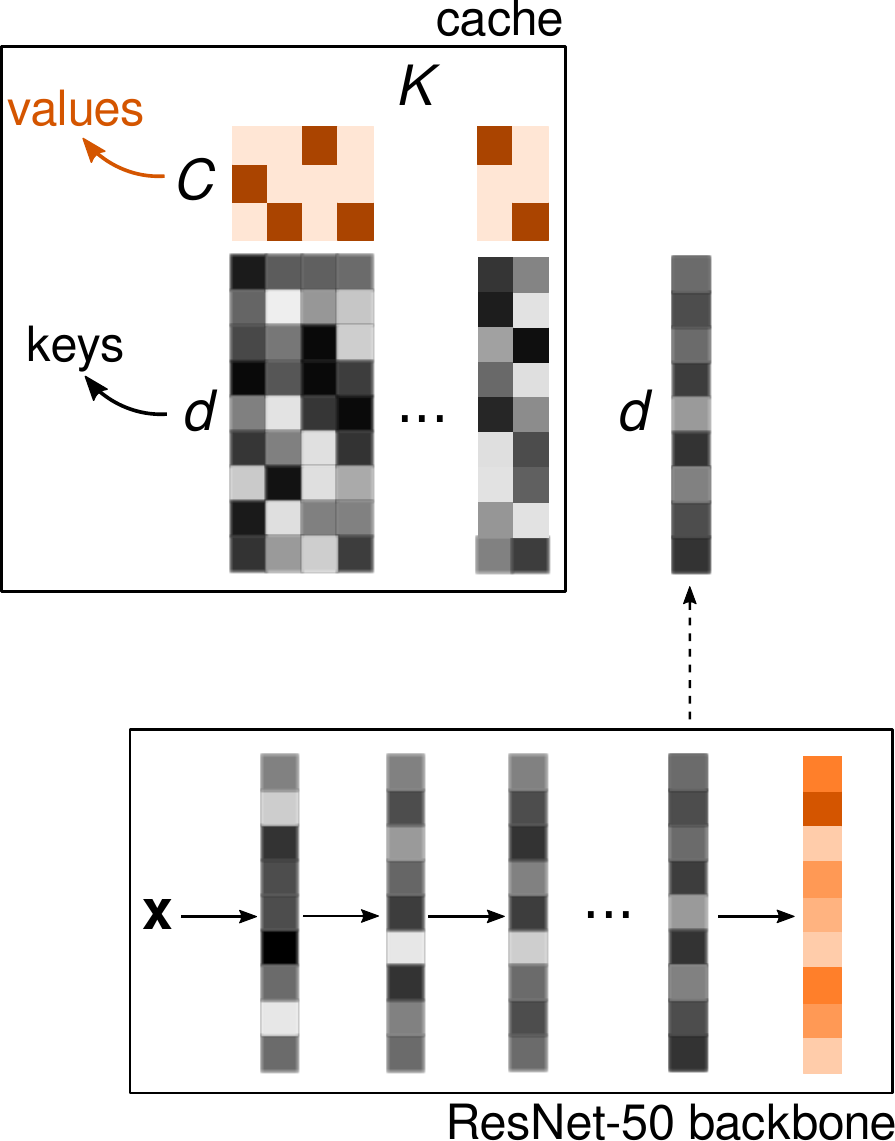}
	\caption{Schematic illustration of the cache model (adapted from \citet{orhan2018}). The key for a new image $\mathbf{x}$ is compared with the keys in the cache. A prediction is made by a linear combination of the values weighted by the similarity to the corresponding keys.} \label{cache_fig}
\end{figure}

Throughout the paper, we use pre-trained ResNet-50 models either on their own or as feature extractors (or ``backbones'') to build cache models that incorporate an explicit episodic memory storing low-dimensional embeddings (or keys) for all images seen during training \citep{orhan2018}. The cache models in this paper are essentially identical to the ``CacheOnly'' models described in \citet{orhan2018}. A schematic diagram of a cache model is shown in Figure~\ref{cache_fig}. 

We used one of the higher layers of a pre-trained ResNet-50 model as an embedding layer. Let $\phi(\mathbf{x})$ denote the $d$-dimensional embedding of an image $\mathbf{x}$ into this layer. The cache is then a key-value dictionary consisting of the keys $\mu_k \equiv \phi(\mathbf{x}_k)$ for each training image $\mathbf{x}_k$ in the dataset and the values are the corresponding class labels represented as one-hot vectors $v_k$ \citep{orhan2018}. We normalized all keys to have unit $l_2$-norm.

When a new test image $\mathbf{x}$ is presented, the similarity between its key and all other keys in the cache is computed through \citep{orhan2018}:
\begin{equation}
\sigma_k(\mathbf{x}) \propto \exp(\theta \phi(\mathbf{x})^\top \mu_k ) \label{cache_sims_eq}
\end{equation} 
A probability distribution over the labels is then obtained by taking an average of the values stored in the cache weighted by the corresponding similarity scores \citep{orhan2018}:
\begin{equation}
p_{\mathrm{cache}}(\mathbf{y}|\mathbf{x}) = \frac{\sum_{k=1}^K\upsilon_k \sigma_k(\mathbf{x})}{\sum_{k=1}^K \sigma_k(\mathbf{x})} \label{predictions_eq}
\end{equation}
where $K$ denotes the number of items stored in the cache. The hyper-parameter $\theta$ in Equation~\ref{cache_sims_eq} acts as an inverse temperature parameter for this distribution, with larger $\theta$ values producing sharper distributions. We optimized $\theta$ only in one of the experimental conditions below (the gray-box adversarial setting) by searching over $9$ uniformly spaced values between $10$ and $90$ and fixed its value for all other conditions.

Because we take all items in the cache into account in Equation~\ref{predictions_eq}, weighted by their similarity to the test item, we call this type of cache a \textit{continuous cache} \citep{grave2016}. An alternative (and more scalable) approach would be to perform a nearest neighbor search in the cache and consider only the most similar items in making predictions \citep{grave2017, dubey2019}. We compare the relative performance of these two approaches below.

For the embedding layer, we considered three choices (in descending order and using the layer names from the \texttt{torchvision.models} implementation of ResNet-50): \texttt{fc}, \texttt{avgpool}, and \texttt{layer4\_bottleneck1\_relu}. \texttt{fc} corresponds to the final softmax layer (we used the post-nonlinearity probabilities, not the logits), \texttt{avgpool} corresponds to the global average pooling layer right before the final layer and \texttt{layer4\_bottleneck1\_relu} is the output of the penultimate bottleneck block of the network. We also explored the use of lower layers as embeddings; however, these layers led to substantially worse clean and adversarial accuracies, hence they were not considered further. \texttt{layer4\_bottleneck1\_relu} is a $7\times7\times2048$-dimensional spatial layer; we applied a global spatial average pooling operation to this layer to reduce its dimensionality. This gave rise to $d=1000$ dimensional keys for \texttt{fc} and $d=2048$ dimensional keys for the other two layers.

To investigate the effect of different feature types on the robustness of the models, we also considered a ResNet-50 model jointly trained on ImageNet and Stylized-ImageNet datasets and then fine-tuned on ImageNet \citep{geirhos2019} (we used the pre-trained model provided by the authors). Following \citet{geirhos2019}, we call this model Shape-ResNet-50. \citet{geirhos2019} argue that Shape-ResNet-50 learns more global, shape-based representations than a standard ImageNet-trained ResNet-50 (which instead relies more heavily on local texture) and produces predictions more in line with human judgments in texture vs. shape cue conflict experiments.

All experiments were conducted on the ImageNet dataset containing approximately $1.28\mathrm{M}$ training images from 1000 classes and $50\mathrm{K}$ validation images \citep{russakovsky2015}. We note that using the full cache (i.e. a continuous cache) was computationally feasible in our experiments at the ImageNet scale. The largest cache we used (of size $1.28\mathrm{M} \times 2048$) takes up $\sim 10.5\mathrm{GB}$ of disk space when stored as a single-precision floating-point array.

\subsection{Perturbations}
Ideally, we want our image recognition models to be robust against both adversarial perturbations and more natural perturbations. This subsection describes the details of the natural and adversarial perturbations considered in this paper.

\subsubsection{Adversarial perturbations}
Our experiments on adversarial perturbations closely followed the experimental settings described in \citet{dubey2019}. In particular, we considered three different threat models: white-box attacks, gray-box attacks, and black-box attacks.

\textbf{White-box attacks:} This is the strongest attack scenario. In this scenario, the attacker has full knowledge of the backbone model and the items stored in the cache.

\textbf{Gray-box attacks:} In this scenario, the attacker has full knowledge of the backbone model, but does not have access to the items stored in the cache. In many cases, this threat scenario is more realistic than the white-box or black-box settings, since the models used as feature extractors are usually publicly available (e.g. pre-trained ImageNet models), but the database of items stored using those features is private. In practice, for the cache models, we implemented the gray-box scenario by first running white-box attacks against the backbone model and then testing the resulting adversarial examples on the cache model.

\textbf{Black-box attacks:} This is the weakest attack scenario where the attacker does not know the backbone model or the items stored in the cache. For the cache models, we implemented the black-box scenario by running white-box attacks against a model different from the model used as the backbone and testing the resulting adversarial examples on the cache model as well as on the backbone itself. In practice, we used an ImageNet-trained ResNet-18 model to generate adversarial examples in this setting (note that we always use a ResNet-50 backbone in our models).

We chose a strong, state-of-the-art, gradient-based attack method called projected gradient descent (PGD) with random starts \citep{madry2017} to generate adversarial examples in all three settings. We used the Foolbox implementation of this attack \citep{rauber2017}, \texttt{RandomStartProjectedGradientDescentAttack},~with the following attack parameters:~\texttt{binary\_search = False},~\texttt{stepsize = 2/225}, \texttt{iterations =	10},~\texttt{random\_start = True}.~We also controlled the total size of the adversarial perturbation as measured by the $l_\infty$-norm of the perturbation normalized by the $l_\infty$-norm of the clean image $\mathbf{x}$: $\epsilon \equiv ||\mathbf{x}_{\mathrm{adv}} - \mathbf{x}||_\infty / ||\mathbf{x}||_\infty$. We considered six different $\epsilon$ values: $0.01$, $0.02$, $0.04$, $0.06$, $0.08$, $0.1$. In general, the attacks are expected to be more successful for larger $\epsilon$ values. 

As recommended by \citet{athalye2018}, we used targeted attacks, where for each validation image we first chose a target class label different from the correct class label for the image and then ran the attack to return an image that was misclassified as belonging to the target class. In cases where the attack was not successful, the original clean image was returned, therefore the model had the same baseline accuracy on such failure cases as on clean images. We ran attacks starting from all validation images, hence the reported accuracies are averages over all validation images. 

\subsubsection{Natural perturbations}
To measure the robustness of image recognition models against natural perturbations, we used the recently introduced ImageNet-C benchmark \citep{hendrycks2019}. ImageNet-C contains 15 different natural perturbations applied to each image in the ImageNet validation set at 5 different severity levels, for a total of $15\times 5\times 50\mathrm{K}=3.75\mathrm{M}$ images. The perturbations in ImageNet-C come in four different categories: 1) \textit{noise} perturbations (Gaussian, shot, and impulse noise), 2) \textit{blur} perturbations (defocus, glass, motion, and zoom blur), 3) \textit{weather} perturbations (snow, frost, fog, and brightness), and 4) \textit{digital} perturbations (contrast, elasticity, pixelation, and JPEG compression). We refer the reader to \citet{hendrycks2019} for further details about the dataset.

To measure the robustness of a model against the perturbations in ImageNet-C, we use the $mCE$ (mean corruption error) measure \citep{hendrycks2019}. A model's $mCE$ is calculated as follows. For each perturbation $c$, we first average the model's classification error over the 5 different severity levels $s$ and divide the result by the average error of a reference classifier (which is taken to be the AlexNet): $CE_c \equiv \langle E_{s,c} \rangle_s / \langle E^{\mathrm{AlexNet}}_{s,c} \rangle_s$. The overall performance on ImageNet-C is then measured by the mean $CE_c$ averaged over the 15 different perturbation types $c$: $mCE\equiv\langle CE_c \rangle_c$. Dividing by the performance of a reference model in calculating $CE_c$ ensures that different perturbations have roughly similar sized contributions to the overall measure \textit{mCE}. Note that smaller \textit{mCE} values indicate more robust classifiers.

\section{Results}
\subsection{Caching improves robustness against adversarial perturbations}
\begin{figure}
	\centering
	\includegraphics[width=0.5\textwidth, trim=0mm 0mm 0mm 0mm, clip]{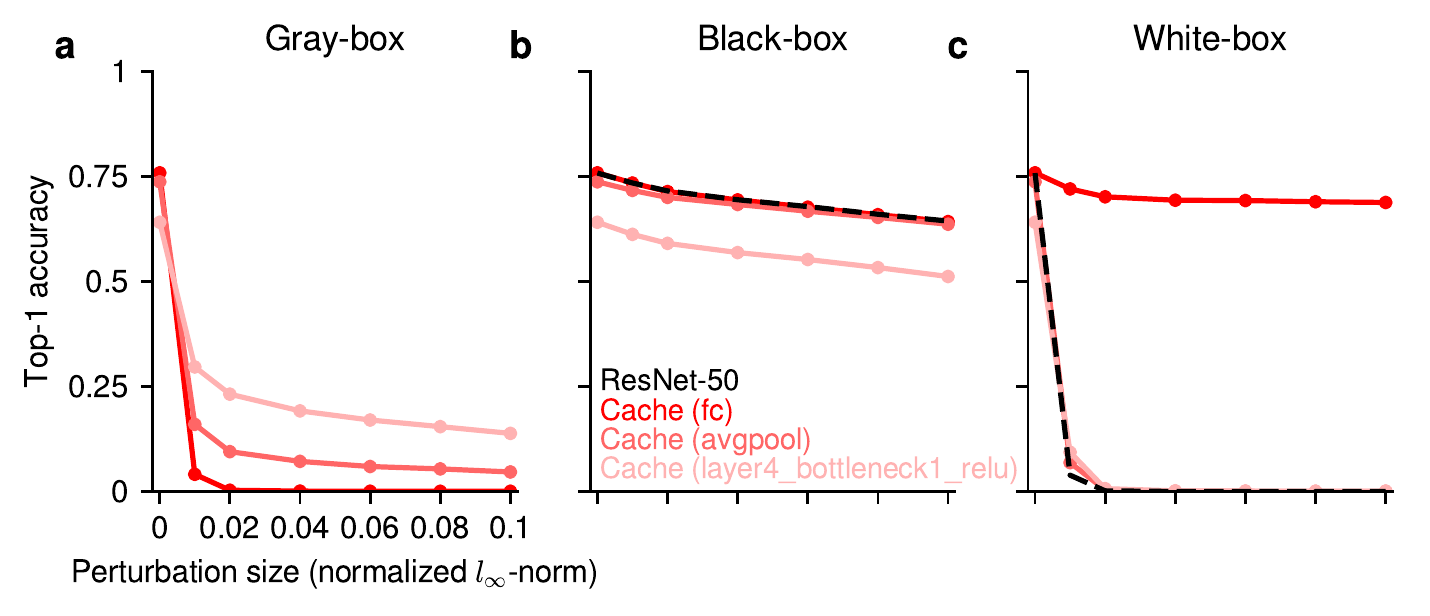}
	\caption{Top-1 accuracy of the ResNet-50 backbone and cache models in the (\textbf{a}) gray-box, (\textbf{b}) black-box and (\textbf{c}) white-box adversarial settings. The $0$ perturbation size corresponds to the clean images. Note that the gray-box setting is meaningful for the cache models only and is not well-defined for the backbone ResNet-50 model.} \label{layerwise_adversarial_fig}
\end{figure}

Figure~\ref{layerwise_adversarial_fig} shows the adversarial accuracy in the gray-box, black-box, and white-box settings for cache models using different layers as embeddings. In the gray-box setting, lower layers showed more robustness at the expense of a reduction in clean accuracy, with the \texttt{layer4\_bottleneck1\_relu} layer achieving the highest gray-box accuracies. 

In the black-box setting, we found that even large perturbation adversarial examples for the ResNet-18 model were not effective adversarial examples for the backbone ResNet-50 model (dashed line) or for the cache models, hence the models largely maintained their performance on clean images with a slight general decrease in accuracy for larger perturbation sizes. 

In the white-box setting, we observed a divergence in behavior between \texttt{fc} and the other layers. The PGD attack was generally unsuccessful against the \texttt{fc} layer cache model, whereas for the other layers it was highly successful even for small perturbation sizes. The softmax non-linearity in \texttt{fc} was crucial for this effect, as it was substantially easier to run successful white-box attacks when the logits were used as keys instead. We thus attribute this effect to gradient obfuscation in the \texttt{fc} layer cache model \citep{athalye2018}, rather than consider it as a real sign of adversarial robustness. Indeed, the gray-box adversarial examples (generated from the backbone ResNet-50 model) were very effective against the \texttt{fc} layer cache model (Figure~\ref{layerwise_adversarial_fig}a).

\begin{figure}
	\centering
	\includegraphics[width=0.5\textwidth, trim=0mm 0mm 0mm 0mm, clip]{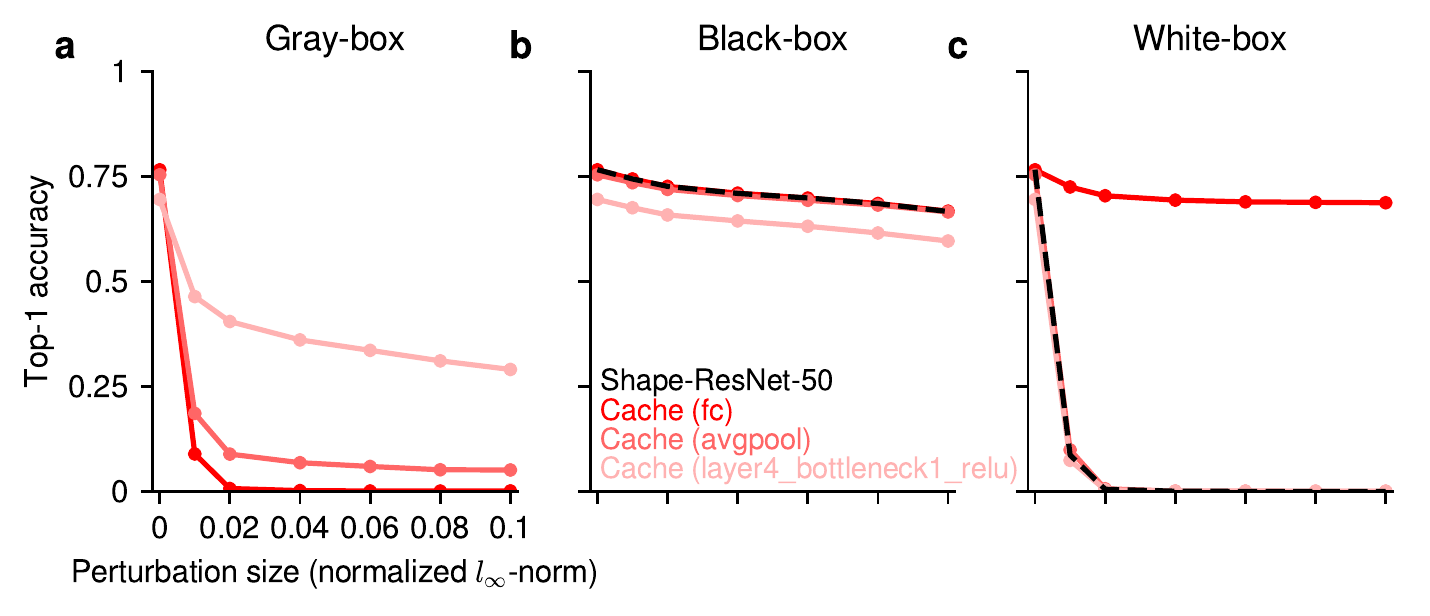}
	\caption{Similar to Figure~\ref{layerwise_adversarial_fig}, but with Shape-ResNet-50 as the backbone.} \label{sin_fig}
\end{figure}

Qualitatively similar results were observed when Shape-ResNet-50 was used as the backbone instead of ResNet-50 (Figure~\ref{sin_fig}). Table~\ref{shape_texture_table} reports the clean and adversarial accuracies for a subset of the conditions.

\begin{table*}[h!]
	\caption{Clean and adversarial accuracies of texture- and shape-based ResNet-50 backbones and cache models. The adversarial accuracies report the results for a standard normalized perturbation size of $\epsilon=0.06$.}
	\label{shape_texture_table}
	\centering
\begin{tabular}{lcccc}
	\toprule
	Model                           & Clean & Gray-box & Black-box & White-box \\
	\midrule
	ResNet-50                       & 0.758 & --       & 0.678     & 0.000     \\
	Cache (\texttt{layer4\_bottleneck1\_relu}, texture) & 0.641 & 0.170    & 0.552     & 0.001     \\
	Shape-ResNet-50                 & 0.766 & --       & 0.699     & 0.000     \\
	Cache (\texttt{layer4\_bottleneck1\_relu}, shape)   & 0.695 & 0.336    & 0.632     & 0.001     \\
	\bottomrule
\end{tabular}
\end{table*}

\subsection{Cache design choices}
In this subsection, we consider the effect of three cache design choices on the clean and adversarial accuracy of cache models: the size and dimensionality of the cache and the retrieval method. 

\citet{dubey2019} recently investigated the adversarial robustness of cache models with very large databases (databases of up to $K=50\mathrm{B}$ items). Scaling up the cache model to very large databases requires making the cache memory as compact as possible and using a fast approximate nearest neighbor algorithm for retrieval from the cache (instead of using a continuous cache). There are at least two different ways of making the cache more compact: one can either reduce the number of items in the cache by clustering them, or alternatively one can reduce the dimensionality of the keys.

\citet{dubey2019} made the keys more compact by reducing the original $2048$-dimensional embeddings to $256$ dimensions (an $8$-fold compression) with online PCA and used a fast 50-nearest neighbor (50-nn) method for retrieval.

\begin{table*}
	\caption{Clean and gray-box adversarial accuracies of different cache models. As in Figure~\ref{cache_type_fig}, only the results for the \texttt{layer4\_bottleneck1\_relu} layer are shown. Colors highlight the {\color{red}retrieval method} (continuous or 50-nn), {\color{blue}cache dimensionality} (full, $4$- or $8$-times reduced), and {\color{magenta}cache size} (full, $4$- or $8$-times reduced).}
	\label{cache_type_table}
	\centering
	\begin{tabular}{lcc}
		\toprule
		Model & Clean & Gray-box ($\epsilon=0.06$) \\
		\midrule
		Cache ({\color{red}continuous}, {\color{blue}full-dims.}, {\color{magenta}full-cache}) & 0.641 & 0.170 \\
		Cache ({\color{red}50-nn}, {\color{blue}full-dims.}, {\color{magenta}full-cache}) & 0.656 & 0.181   \\
		Cache ({\color{red}50-nn}, {\color{blue}full-dims.}, {\color{magenta}$1/4$-cache}) & 0.626 & 0.151   \\
		Cache ({\color{red}50-nn}, {\color{blue}full-dims.}, {\color{magenta}$1/8$-cache}) & 0.612 & 0.143   \\
		Cache ({\color{red}50-nn}, {\color{blue}$1/4$-dims.}, {\color{magenta}full-cache}) & 0.516 & 0.109   \\
		Cache ({\color{red}50-nn}, {\color{blue}$1/8$-dims.}, {\color{magenta}full-cache}) & 0.498 & 0.103   \\
		\bottomrule
	\end{tabular}
\end{table*}

\begin{figure}
	\centering
	\includegraphics[width=0.5\textwidth,trim=0mm 0mm 0mm 0mm,clip]{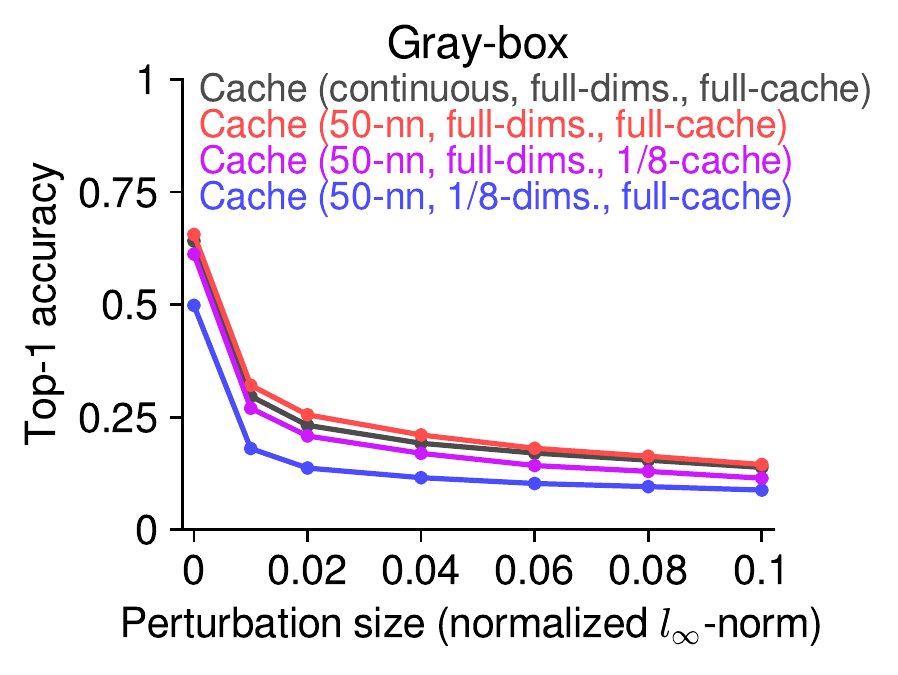}
	\caption{The effects of three cache design choices on the clean and adversarial accuracy in the gray-box setting. The results shown here are for the \texttt{layer4\_bottleneck1\_relu} layer. Similar results were observed for other layers.} \label{cache_type_fig}
\end{figure}

In our experiments, replacing the continuous cache with a 50-nn retrieval method did not have an adverse effect on adversarial and clean accuracies (Figure~\ref{cache_type_fig} and Table~\ref{cache_type_table}). This suggests that the continuous cache can be safely replaced with an efficient nearest neighbor algorithm to scale up the cache size without much effect on the model accuracy. 

On the other hand, reducing the dimensionality of the keys from $2048$ to $256$ using online PCA over the training data resulted in a substantial drop in both clean and adversarial accuracies (Figure~\ref{cache_type_fig} and Table~\ref{cache_type_table}). Even a $4$-fold reduction to $512$ dimensions resulted in a large drop in accuracy. This implies that the higher layers of the backbone used for caching are not very compressible and drastic dimensionality reduction measures should be avoided to prevent a substantial decrease in accuracy.

Reducing the cache size by the same amount ($4$-fold or $8$-fold compression) by clustering the items in the cache with a mini-batch \textit{k}-means algorithm resulted in a significantly smaller decrease in accuracy (Figure~\ref{cache_type_fig} and Table~\ref{cache_type_table}): for example, an $8$-fold reduction in dimensionality led to a clean accuracy of $49.8\%$, whereas an $8$-fold reduction in the cache size instead resulted in a clean accuracy of $61.2\%$. This suggests that the cluster structure in the keys is much more prominent than the linear correlation between the dimensions. Therefore, to make the cache more compact, given a choice between reducing the dimensionality vs. reducing the number of items by the same amount, it is preferable to choose the second option for better accuracy.

\subsection{Caching does not improve robustness against natural perturbations}
We have seen that caching can improve robustness against gray-box adversarial perturbations. Does it also improve robustness against more natural perturbations? Table~\ref{imagenet_c} shows that the answer is no. On ImageNet-C, the backbone ResNet-50 model yields an $mCE$ of 0.764. The best cache model obtained approximately the same $mCE$ score. We suggest that this is because caching improves robustness only against small-norm perturbations, whereas natural perturbations in ImageNet-C are typically much larger. Even the smallest size perturbations in ImageNet-C are clearly visible to the eye \citep{hendrycks2019} and we calculated that even these smallest size perturbations have an average normalized $l_\infty$-norm of $\epsilon \approx 1$ for all perturbation types, compared to the largest adversarial perturbation size of $\epsilon = 0.1$ considered in this paper. This result is also consistent with a similar observation made by \citet{gu2019} suggesting that perturbations occurring between neighboring frames in natural videos are much larger in magnitude than adversarial perturbations. We conjecture that robustness against such large perturbations cannot be achieved with test-time only interventions such as caching and requires learning more robust backbone features in the first place.

\subsection{Using shape-based features in the cache improves both adversarial and natural robustness}
To investigate the effect of different kinds of features in the cache, we repeated our experiments using cache models with Shape-ResNet-50 as the backbone (see \textit{Methods} for further details about Shape-ResNet-50). It has been argued that Shape-ResNet-50 learns more global, shape-based representations than a standard ImageNet-trained ResNet-50 and it has already been shown to improve robustness on the ImageNet-C benchmark \citep{geirhos2019}. We confirm this improvement (Table~\ref{imagenet_c}; ResNet-50 vs. Shape-ResNet-50) and find that caching with Shape-ResNet-50 leads to roughly the same $mCE$ as the backbone Shape-ResNet-50 itself. 

\begin{table*}
	\caption{ImageNet-C results. The numbers indicate corruption errors ($CE$) for specific corruption types and the mean $CE$ scores as percentages. More robust models correspond to smaller numbers. For the cache models, we only show the results for the best models (the \texttt{fc} cache model in both cases). Colors represent noise, {\color{red}blur}, {\color{blue}weather} and {\color{magenta}digital} perturbations.}
	\label{imagenet_c}
	\centering
	\setlength{\tabcolsep}{0.175\tabcolsep}
	\begin{tabular}{lc|ccccccccccccccc}
		\toprule
		Model & {\textit{mCE}} & {\tiny Gauss} & {\tiny Shot} & {\tiny Impul.} & {\tiny {\color{red}Defoc.}} & {\tiny {\color{red}Glass}} & {\tiny {\color{red}Motion}} & {\tiny {\color{red}Zoom}} & {\tiny {\color{blue}Snow}} & {\tiny {\color{blue}Frost}} & {\tiny {\color{blue}Fog}} & {\tiny {\color{blue}Bright}} & {\tiny {\color{magenta}Contr.}} & {\tiny {\color{magenta}Elastic}} & {\tiny {\color{magenta}Pixel}} & {\tiny {\color{magenta}JPEG}} \\
		\midrule
		ResNet-50       & 76.4 & 78 & 80 & 80 & 75 & 89 & 78 & 80 & 78 & 75 & 67 & 57 & 72 & 86 & 77 & 76 \\
		Cache (texture) & 76.4 & 78 & 79 & 80 & 75 & 89 & 78 & 80 & 78 & 75 & 67 & 57 & 72 & 86 & 77 & 76 \\
		Shape-ResNet-50 & 73.5 & 74 & 75 & 75 & 72 & 86 & 74 & 80 & 75 & 73 & 67 & 55 & 68 & 81 & 75 & 72 \\
		Cache (shape)   & 73.5 & 74 & 75 & 75 & 72 & 86 & 74 & 80 & 75 & 73 & 67 & 55 & 68 & 81 & 75 & 72 \\
		\bottomrule
	\end{tabular}
\end{table*}

\begin{figure}
	\centering
	\includegraphics[width=0.5\textwidth, trim=0mm 0mm 0mm 0mm, clip]{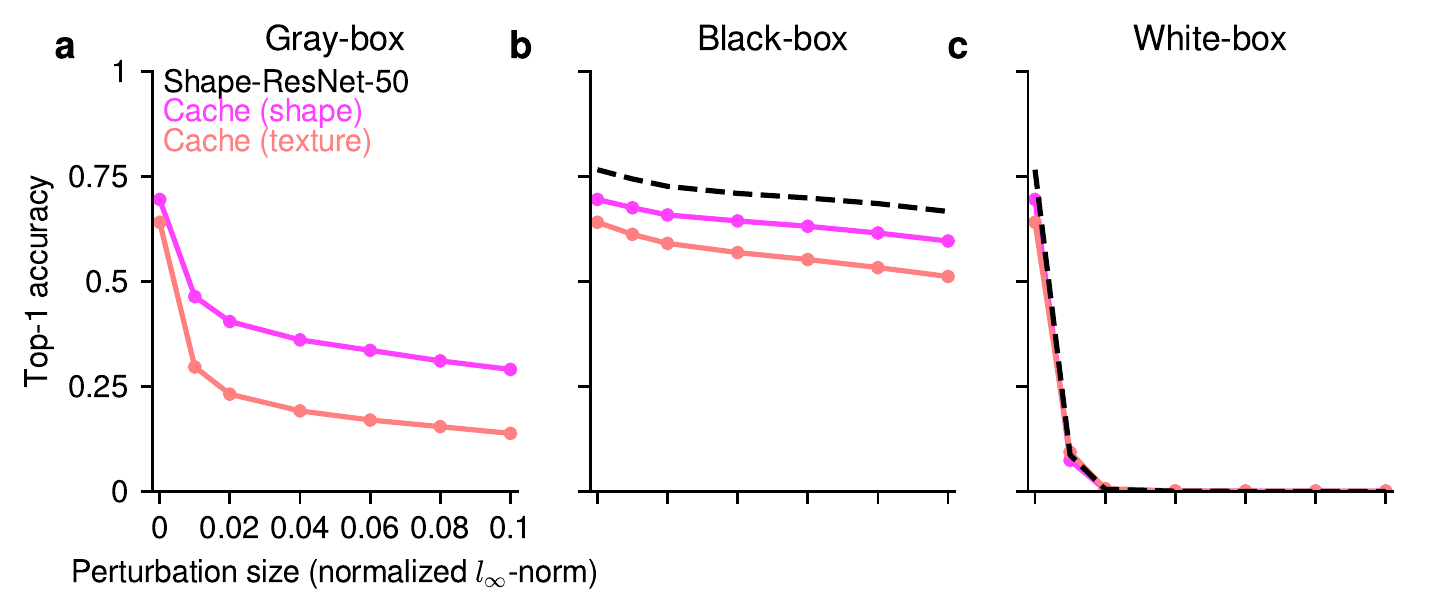}
	\caption{The effect of using Shape-ResNet-50 (shape) vs. ResNet-50 (texture) derived features in the cache on clean and adversarial accuracies in the (\textbf{a}) gray-box, (\textbf{b}) black-box and (\textbf{c}) white-box settings. The results shown here are for the \texttt{layer4\_bottleneck1\_relu} layer.} \label{shape_vs_texture_fig}
\end{figure}

Remarkably, however, when used in conjunction with caching, these Shape-ResNet-50 features also substantially improved the adversarial robustness of the cache models in the gray-box and black-box settings, compared to the ImageNet-trained ResNet-50 features. Figure~\ref{shape_vs_texture_fig} illustrates this for the \texttt{layer4\_bottleneck1\_relu} cache model. This effect was more prominent for earlier layers.

Shape-based features, however, did not improve the adversarial robustness in the white-box setting, neither for the backbone model nor for the cache models (Table~\ref{shape_texture_table}). This suggests that eliminating the heavy texture bias of DNNs does not necessarily eliminate the existence of adversarial examples for these models. The opposite, however, seems to be true: that is, adversarially robust models do not display a texture bias; instead they seem to be much more shape-biased, similar to humans \citep{zhang2019}.

\section{Discussion}
In this paper, we have shown that a combination of two basic ideas motivated by the cognitive psychology of human vision, an explicit cache-based episodic memory and a shape bias, improves the robustness of image recognition models against both natural and adversarial perturbations at the ImageNet scale. Caching alone improves (gray-box) adversarial robustness only, whereas a shape bias improves natural robustness only. In combination, they improve both, with a synergistic effect in adversarial robustness (Table~\ref{summary}).

\begin{table}[]
	\caption{Summary of our main results. This table is a distilled version of Tables~\ref{shape_texture_table} and \ref{imagenet_c}. In each cell, the first number represents the adversarial accuracy (gray-box accuracy for cache models and white-box accuracy for cacheless models, both with $\epsilon=0.06$); the second number represents the \textit{mCE} score. Note that better models have higher accuracy and lower \textit{mCE} score. Starting from a baseline model with no cache and no shape bias (bottom right), adding a cache memory (bottom left) only improves adversarial accuracy; adding a shape bias (top right) only improves natural robustness; adding both (top left) improves both natural and adversarial robustness with a synergistic improvement in the latter.}
	\label{summary}
	\centering
	\renewcommand{\arraystretch}{1.5}
	\begin{tabular}{l|ll|ll}
		& \multicolumn{2}{c|}{Cache +} & \multicolumn{2}{c}{Cache -} \\ \hline
		Shape bias + & \textbf{33.6\%} & \textbf{73.5} & 0.0\%          & \textbf{73.5} \\ \hline
		Shape bias - & 17.0\%          & 76.4          & 0.0\%          & 76.4        
	\end{tabular}
\end{table}

Why does caching improve adversarial robustness? \citet{orhan2018} suggested that caching acts as a regularizer. More specifically it was shown in \citet{orhan2018} that caching significantly reduces the Jacobian norm at test points, which could explain its improved robustness against small-norm perturbations such as adversarial attacks. However, since Jacobian norm only measures local sensitivity, this does not guarantee improved robustness against larger perturbations, such as the natural perturbations in the ImageNet-C benchmark and indeed we have shown that caching, by itself, does not provide any improvement against such perturbations. 

It should also be emphasized that caching improves adversarial robustness only under certain threat models. We have provided evidence for improved robustness in the gray-box setting only, \citet{zhao2018} and \citet{dubey2019} also provide evidence for improved robustness in the black-box setting (\citet{orhan2018} reports evidence for improved robustness through caching in the white-box setting in CIFAR-10 models, however it is likely that such robustness improvements are much easier to achieve in CIFAR-10 models than in ImageNet models). The results in \citet{dubey2019} are particularly encouraging, since they suggest that the caching approach can scale up in the gray-box and black-box attack scenarios in the sense that larger cache sizes lead to more robust models. On the other hand, neither of these two earlier works, nor our own results point to any substantial improvement in adversarial robustness in the white-box setting at the ImageNet scale. The white-box setting is the most challenging setting for an adversarial defense. Theoretical results suggest that in terms of sample complexity, robustness in the white-box setting may be fundamentally more difficult than achieving high generalization accuracy in the standard sense \citep{schmidt2018, gilmer2018} and it seems unlikely that it can be feasibly achieved via test-time only interventions such as caching.

Why does a shape bias improve natural robustness? Natural perturbations modeled in ImageNet-C typically corrupt local information, but preserve global information such as shape. Therefore a model that can integrate information more effectively over long distances, for example by computing a global shape representation is expected to be more robust to such natural perturbations. In Shape-ResNet-50 \citep{geirhos2019}, this was achieved by removing the local cues to class label in the training data. In principle, a similar effect can be achieved through architectural inductive biases as well. For example, \citet{hendrycks2019} showed that the so-called feature aggregating architectures such as the ResNeXt architecture \citep{xie2017} are substantially more robust to natural perturbations than the ResNet architecture, suggesting that they are more effective at integrating local information into global representations. However, it remains to be seen whether such feature aggregating architectures accomplish this by computing a \textit{shape} representation.

In this work, we have also provided important insights into several cache design choices. Scaling up the cache models to datasets substantially larger than ImageNet would require making the cache as compact as possible. Our results suggest that other things being equal, this should be done by clustering the keys rather than by reducing their dimensionality. For very large datasets, the continuous cache retrieval method that uses the entire cache in making predictions (Equations~\ref{cache_sims_eq} and \ref{predictions_eq}) can be safely replaced with an efficient $k$-nearest neighbor retrieval algorithm, e.g. Faiss \citep{johnson2017}, without incurring a large cost in accuracy. Our results also highlight the importance of the backbone choice (for example, Shape-ResNet-50 vs. ResNet-50): in general, starting from a more robust backbone should make the cache more effective against both natural and adversarial perturbations.

In future work, we are interested in applications of naturally and adversarially robust features in few-shot recognition tasks and in modeling neural and behavioral data from humans and monkeys \citep{schrimpf2018}.

{\small
	\bibliographystyle{iclr2020_conference}
	\bibliography{OrhanLake2020_cvpr}
}
\end{document}